\def\year{2019}\relax
\documentclass[letterpaper]{article}
\usepackage{aaai19} 
\usepackage{times}  
\usepackage{helvet} 
\usepackage{courier} 
\usepackage[hyphens]{url} 
\usepackage{graphicx} 
\usepackage{graphicx}  
\usepackage{tikz}
\usetikzlibrary{positioning}
\usetikzlibrary{shapes.geometric}
\usepackage{bm} 
\usepackage{subcaption}
\usepackage{booktabs}
\usepackage{array}
\newcolumntype{P}[1]{>{\raggedright\arraybackslash}p{#1}}

\usepackage{amsmath}
\DeclareMathOperator*{\argmax}{arg\,max}

\usepackage{todonotes}
\nocopyright

\title{Language-guided Adaptive Perception with Hierarchical Symbolic Representations for Mobile Manipulators}

\author{Ethan Fahnestock, Siddharth Patki, Thomas M. Howard\\
University of Rochester\\
Hajim School of Engineering and Applied Sciences\\
Rochester, NY 14627\\
efahnest@u.rochester.edu\\
}
 
\begin{document}

\maketitle
\begin{abstract}
Language is an effective medium for bi-directional communication in human-robot teams. 
To infer the meaning of many instructions, robots need to construct a model of their surroundings that describe the spatial, semantic, and metric properties of objects from observations and prior information about the environment.
Recent algorithms condition the expression of object detectors in a robot's perception pipeline on language to generate a minimal representation of the environment necessary to efficiently determine the meaning of the instruction. 
We expand on this work by introducing the ability to express hierarchies between detectors. 
This assists in the development of environment models suitable for more sophisticated tasks that may require modeling of kinematics, dynamics, and/or affordances between objects.
To achieve this, a novel extension of symbolic representations for language-guided adaptive perception is proposed that reasons over single-layer object detector hierarchies.
Differences in perception performance and environment representations between adaptive perception and a suitable exhaustive baseline are explored through physical experiments on a mobile manipulator. 
\end{abstract}

\section{Introduction}

Natural language provides a powerful medium for communication with artificial agents.
By enabling collaborative robots to understand and communicate like other humans, the cost of effectively integrating robots into human-robot teams is lowered significantly, since operators will not be required to learn engineered interfaces that may be less accessible to novice users.

The proliferation and improvements in natural language voice assistants has demonstrated that artificial agents are capable of completing simple tasks, such as adding reminders, checking the weather, or making calls. 
However, contemporary voice assistants do not consider models of their physical surroundings when engaging in dialogue. Such assistants will not be able to answer questions like ``what is the color of the box on my left'' because they do not construct or utilize a representation of the local environment when interpreting the meaning of the sentence. 
For language to be an effective tool for human-robot teams, robots must understand instructions in the context of its surroundings.
Robots build and maintain an internal representation of the outside world to plan and execute useful behaviors. This world can be used by robots to understand and act on natural language instructions. 
This dependence on the world model means that a robot's ability to understand and act on natural language is limited by their representation of the world. 
To understand and attempt a diverse set of tasks, a robot needs a perception pipeline capable of detecting, classifying, and modeling a diverse set of objects.

For example, if the robot only models doors as static landmarks with a pose and bounding box, a human teammate cannot ask the robot to ``open the door by the handle'', since there is no representation of the door's handle. 
\begin{figure}[t]
	\centering
	\begin{subfigure}[b]{0.44\columnwidth}
		\centering
		\includegraphics[width=\textwidth]{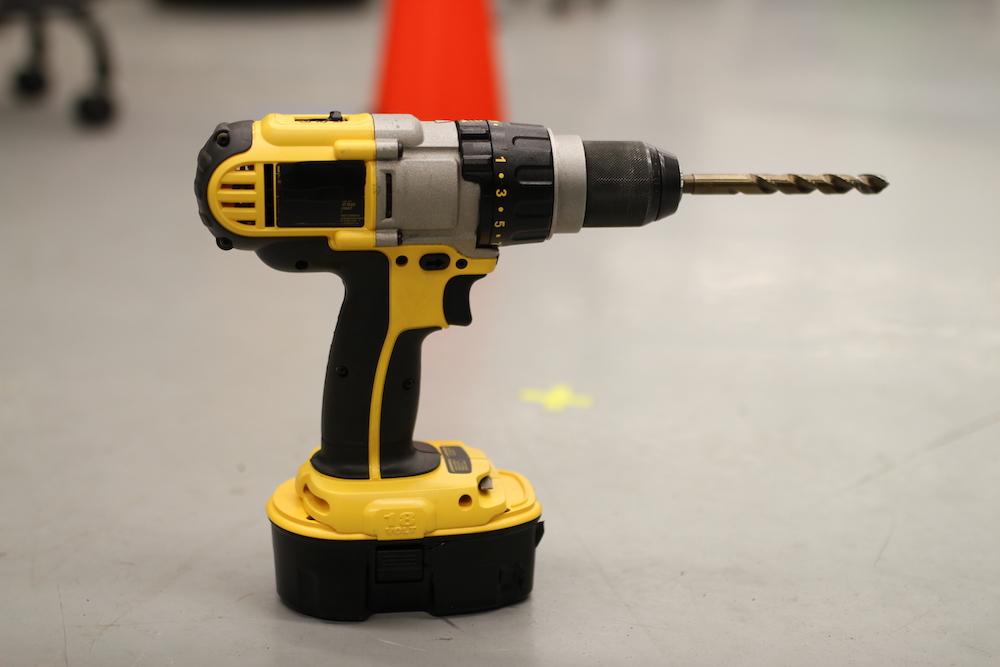}
	\end{subfigure}
	\begin{subfigure}[b]{0.44\columnwidth}
		\centering
		\includegraphics[width=\textwidth]{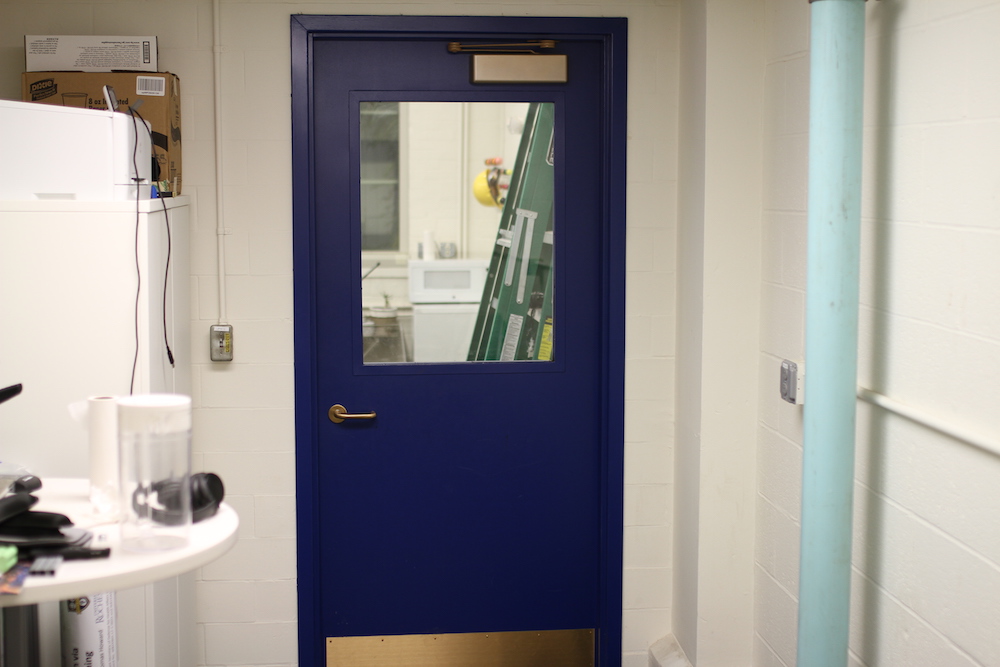}
	\end{subfigure}
	\begin{subfigure}[b]{0.44\columnwidth}
		\centering
		\includegraphics[width=\textwidth]{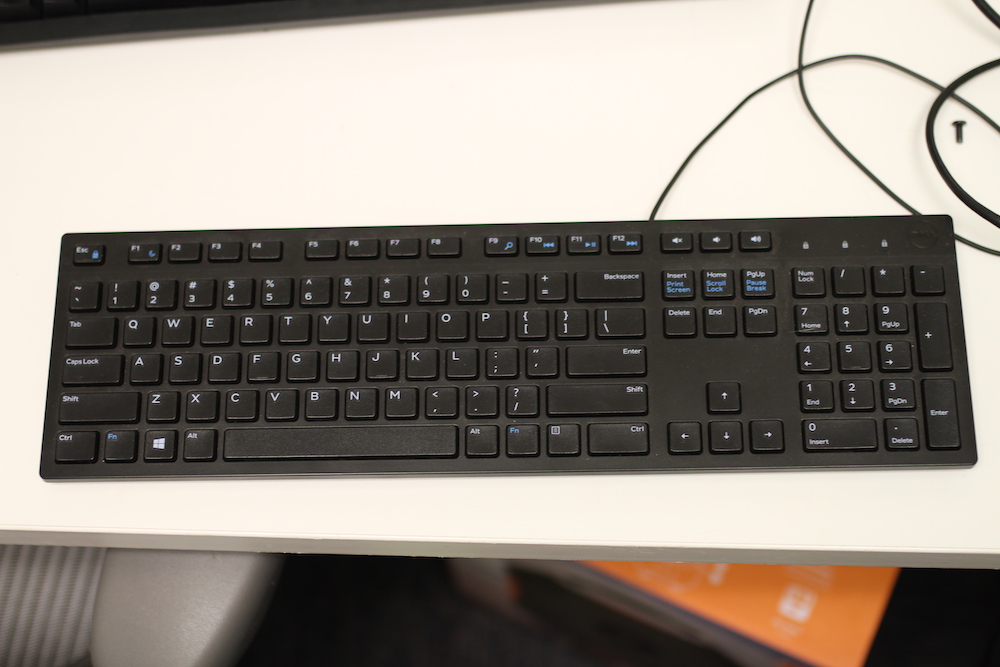}
	\end{subfigure}
	\begin{subfigure}[b]{0.44\columnwidth}
		\centering
		\includegraphics[width=\textwidth]{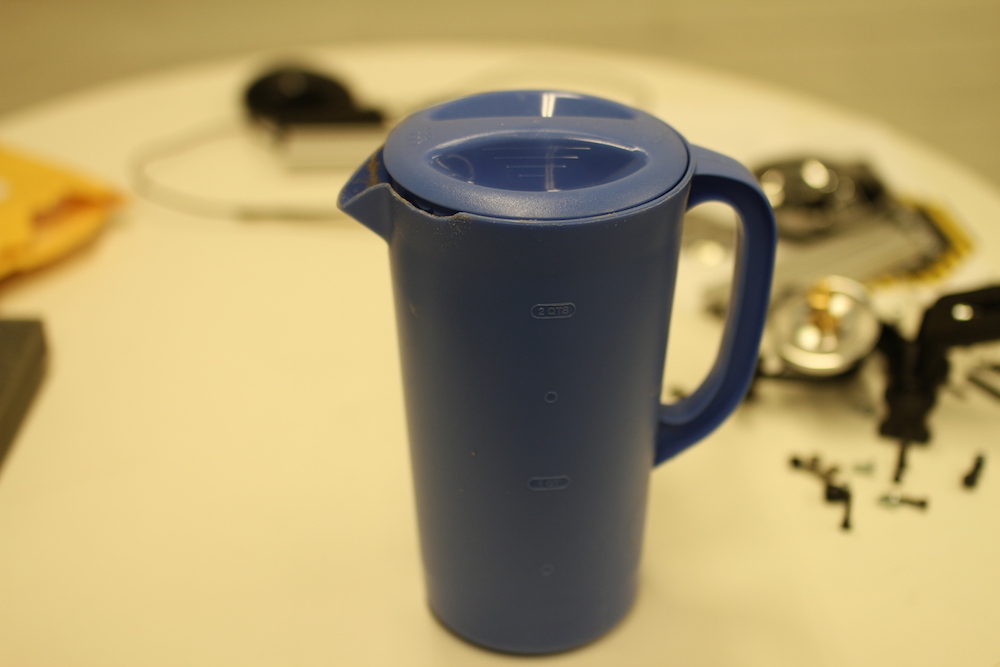}
	\end{subfigure}

  \caption{An example of objects with natural hierarchies. The drill has a handle, battery, trigger, and drill bit. The door has a handle and a window. The keyboard has a set of keys, a wire, and a frame. A robot tasked with picking up the drill by its handle may require the ability to build hierarchical representations of the drill to complete its task. We propose a language-guided perception pipeline that infers and selectively represents hierarchies of object detectors to better construct minimally viable worlds to ground natural language and perform planning in.}
  \label{fig:motivation}
\end{figure}
This limitation does not only apply to the presence of objects in the world model, but the level of detail, or fidelity, to which they are represented. 
In world models where all objects are assumed to be static rigid bodies, sophisticated motion planning algorithms may fail to find solutions to more complex tasks that involve navigation and/or manipulation of articulated and/or dynamic objects.  
Conversely, overly rich representations of the environment that model dynamics and affordances that are not pertinent to a robot's task consume computational resources that would be better used in other parts of the robot's intelligence architecture.  
Algorithms that enable a robot to fluidly move across this spectrum of environment representation fidelity will enable more efficient execution of a diverse range of tasks.

One aspect of environment representation fidelity that may need to be represented are the hierarchical relationships between objects.
Consider the objects depicted in Figure \ref{fig:motivation}. 
Each of these objects can be decomposed into their constituents, which may be vital depending on the task. 
A robot may be tasked with removing the drill bit from the drill, opening the door using the door handle, pressing the enter key, or or to pickup the pitcher by its handle. 
Approaching the decomposition and modeling of objects and their components enables a robot to approach these tasks, but at the cost of adding complexity to the perception pipeline.
Take the drill as an example. If the robot is tasked with using the drill as a tool, the relationship the drill object has with its drill bit, battery, and trigger are critical to complete the task successfully. If instead the robot is simply asked to transport the drill, these relationships are not important to the task and should not be represented in the world model. 

The cost of building hierarchical world models is three-fold. 
First, processing observations becomes more computationally expensive as classification, segmentation, and modeling of objects expands to represent the constituents of each object.
Second, the symbolic representation the robot uses to ground natural language increases in complexity. 
If language understanding is approached as a search problem, increasing the world complexity necessarily increases the size of the search space. 
Thirdly, simulating the environment in which the robot must plan its actions becomes more computationally expensive, which can limit the speeds at which robots can navigate safely through complex terrain.

Previous work \cite{patki18a,patki2019a} has demonstrated that conditioning the active detectors in a perception pipeline on natural language (termed adaptive perception, or AP) to generate minimally viable worlds to understand the provided command increases the efficiency and scalability of natural language grounding.
This work has advanced our understanding of how to simplify a robot's world model from language, but motivates its simplification of the world model based on the needs of language grounding alone. 
In this paper, we take a first step towards re-scoping adaptive perception to generate minimally sufficient worlds that satisfy the needs of planning and execution along with language grounding.
To take this step we extend adaptive perception to reason over hierarchies of object detectors, allowing the selective representation of hierarchical relationships in perception pipelines conditioned on natural language.

We demonstrate adaptive perception's ability to infer hierarchies implicit in the given command needed to complete a task. 
Experiments are performed on a Clearpath Robotics Husky A200 unmanned ground vehicle with an attached manipulator.  

\section{Related Work}
Recent developments in grounded language communication for human-robot interaction have enabled robots to more fluidly understand and generate natural language across a diverse set of applications and domains.   
Probabilistic models for natural language symbol grounding \cite{tellex11a} showed the ability to construct graphical models that learn the association between parts of speech and objects and trajectories that a robot can perform.  
The relationship between action verbs and the perceivable state changes they represent has been explored by \cite{gao16}. 
Other approaches learned parsing of natural language into a ``robot control language'' \cite{matuszek12a} that allowed for the completion of language dictated tasks in world models not available at the time of parsing.
Models such as those presented in \cite{matuszek12} have demonstrated that perception and language models can be learned jointly to achieve attribute grounding. 
Natural language was also shown to be an effective sensor to assist in learning motion models and affordances when combined with vision sensors in \cite{daniele17a}. 
Additional work has focused on improving the structures used for grounding using human-robot dialog \cite{thomason:icra19}.

Task-specific perception pipelines allow robots to efficiently extract only task-relevant information from the environment. An example of an application of this approach is described in \cite{hudson12a}. We expand this approach by learning to extract perceptual requirements from natural language. 
Our work is most similar to \cite{patki18a}, which approach creating minimal viable world models by introducing the idea of adaptive perception, which infers the subset of detectors required to build worlds for language understanding.
Adaptive perception is further expanded in \cite{patki2019a} through the introduction of observation filtering, which filters observations considered for grounding to spatial regions specified in the instruction, such as "in the kitchen" or ``in the hallway''. 
The problem of reasoning over object hierarchies is not explicitly explored in these works. 
In this paper, we present a novel extension of the symbolic representations for adaptive perception that allows the robot to interpret objects differently based on their expected or planned interactions. Specifically, we introduce symbols that can selectively represent hierarchies between object detectors.

In the following section we review the mathematical framework of adaptive perception, new symbolic representations for object detectors, and the system architecture of detector and behavior inference. This discussion follows with a description of the experimental setup, a presentation of experimental results, and discussion of observations made from these experiments. The paper concludes with a review of the contributions and a description of opportunities for further investigation.   

\section{Technical Approach}

We frame language understanding as a symbol grounding problem \cite{harnad90} where we find the most likely set of groundings $\mathbf{\Gamma} = \{\gamma_1, \gamma_2,..,\gamma_n\}$ given a natural language instruction $\mathbf{\Lambda} = \{\lambda_1, \lambda_2,..,\lambda_n\}$ and the world model $\Upsilon$. This can be formalized as an inference problem:
\begin{equation}
	\mathbf{\Gamma^*} = \argmax_{\gamma_1...\gamma_n \in \mathbf{\Gamma}} p(\mathbf{\Gamma} | \mathbf{\Lambda}, \Upsilon)
	\label{eq:grounding}
\end{equation}

As the symbol space of groundings becomes large, inference in the above equation becomes computationally intractable. 
To address this problem, we apply Distributed Correspondence Graphs (DCGs) \cite{howard14,paul2018efficient}, which assume conditional independence across phrases in the constituency parse $\tau(\mathbf{\Lambda})$. 
DCG's introduce correspondence variables $\phi_{ij}$ which associate the phrase $\lambda_i$ with the grounding constituent $\gamma_{ij}$. Adapting Equation \ref{eq:grounding} to perform inference over the set of correspondence variables under the conditional independence assumptions results in the following:
\begin{equation}
	\mathbf{\Phi^*} = \argmax_{\phi_{ij} \in \mathbf{\Phi}} \prod_{i=1}^{|\mathcal{N}|} \prod_{j=1}^{|\mathbf{\Gamma}|} p\left(\phi_{ij} | \gamma_{ij}, \lambda_i, \mathbf{\Phi_{c_i}} \Upsilon\right)
	\label{eq:dcg-prob}
\end{equation}

Here, $\mathbf{\Phi}$ is the set of correspondence variables, $\mathcal{N}$ is the set of all phrases in the constituency parse $\tau(\mathbf{\Lambda})$, $\mathbf{\Gamma}$ is the set of all grounding symbols, and $\mathbf{\Phi_{c_i}}$ are the child correspondence variables of phrase $\lambda_i$. The conditional probabilities in Equation \ref{eq:dcg-prob} are represented by a function $\Psi$ learned using a log-linear model trained on a corpus of annotated examples.

\begin{equation}
	\mathbf{\Phi^*} = \argmax_{\phi_{ij} \in \mathbf{\Phi}} \prod_{i=1}^{|\mathcal{N}|} \prod_{j=1}^{|\mathbf{\Gamma}|} \Psi\left(\phi_{ij}, \gamma_{ij}, \lambda_i, \mathbf{\Phi_{c_i}} \Upsilon\right)
	\label{eq:dcg}
\end{equation}

Equations \ref{eq:dcg-prob} and \ref{eq:dcg} condition their inference on the world model $\Upsilon$. Adaptive perception \cite{patki18a} proposes reducing this world model to a minimal but sufficient world $\Upsilon^*$ for natural language understanding. This is done by inferring a subset of perceptual classifiers $\mathbf{P^*} \in \mathbf{P}$ required for a given task.

\begin{equation}
	\mathbf{P^*} = f(\mathbf{P}, \mathbf{\Lambda})
	\label{eq:ap-abstract}
\end{equation}

This subset of perception symbols is then used to construct a perception pipeline which generates a task-specific compact world model $\Upsilon^*$ by processing observations.

\begin{equation}
	\Upsilon^* = f_{ap}(\mathbf{P^*}, z_{1:t})
	\label{eq:ap}
\end{equation}

In \cite{patki18a}, minimally viable worlds are defined as worlds containing objects required for language grounding. Equation \ref{eq:ap-abstract} is approximated by using DCG to infer a set of perception symbols $\Gamma^{P}$ from the natural language instruction $\mathbf{\Lambda}$. The perception symbols do not relate to physical objects and thus DCG does not require the world model $\Upsilon$ for inference. This is reflected in the following equation. 

\begin{equation}
	\mathbf{\Phi^*} = \argmax_{\phi_{ij} \in \mathbf{\Phi}} \prod_{i=1}^{|\mathcal{N}|} \prod_{j=1}^{|\mathbf{\Gamma^P}|} \Psi\left(\phi_{ij}, \gamma_{ij}, \lambda_i, \mathbf{\Phi_{c_i}} \right)
\end{equation}

In \cite{patki18a} the set of perception detector symbols is defined as the union of a set of independent $\Gamma^{ID}$ and conditionally dependant $\Gamma^{CD}$ symbols. The independent detector symbols represent independent detectors in the following categories: color, geometry, semantic labels, bounding boxes, spatial relationships, and pose. Conditionally dependant symbols form pairs over independent symbols in specific combinations of independent categories. Specifics on these symbols can be found in \cite{patki18a}.

A limitation of the original formulation is that the space of detectors $\Gamma^{P}$ only considered objects of unique semantic labels (e.g. door, window, box) or basic object properties (e.g., color, pose, and geometry) and only expressed those objects as static entities deprived of affordances in the world model. 
If we consider more complicated tasks beyond simple grasping \cite{patki18a} or navigation \cite{patki2019a}, we will encounter scenarios where the fidelity of the world representation needed by the behavior planner and executor nodes will be richer than the minimally viable worlds $\Upsilon^*$ needed for inferring symbolic behaviors.

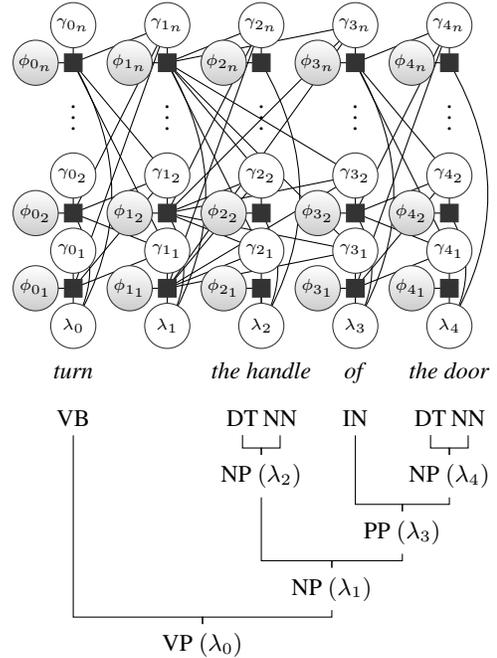
\begin{figure}[ht]
	\centering
	\begin{tikzpicture}[textnode/.style={anchor=mid,font=\tiny},nodeknown/.style={circle,draw=black!80,fill=white,minimum size=6mm,font=\tiny},nodeunknown/.style={circle,draw=black!80,fill=black!10,minimum size=6mm,font=\tiny,top color=white,bottom color=black!20},factor/.style={rectangle,draw=black!80,fill=black!80,minimum size=2mm,font=\tiny,text=white}]
\draw[-] (0,0.5) to (0,1);
\draw[-] (0,0.5) to [bend right=30] (0,2);
\draw[-] (0,0.5) to [bend right=30] (0,4);
\draw[-] (-0.5,1) to (0,1);
\draw[-] (-0.5,2) to (0,2);
\draw[-] (-0.5,4) to (0,4);
\draw[-] (0,1.5) to (0,1);
\draw[-] (0,2.5) to (0,2);
\draw[-] (0,4.5) to (0,4);
\draw[-] (1.25,1.5) to (0,1);
\draw[-] (1.25,2.5) to (0,1);
\draw[-] (1.25,4.5) to (0,1);
\draw[-] (1.25,1.5) to (0,2);
\draw[-] (1.25,2.5) to (0,2);
\draw[-] (1.25,4.5) to (0,2);
\draw[-] (1.25,1.5) to (0,4);
\draw[-] (1.25,2.5) to (0,4);
\draw[-] (1.25,4.5) to (0,4);
\draw[-] (2.5,1.5) to (1.25,1);
\draw[-] (2.5,2.5) to (1.25,1);
\draw[-] (2.5,4.5) to (1.25,1);
\draw[-] (2.5,1.5) to (1.25,2);
\draw[-] (2.5,2.5) to (1.25,2);
\draw[-] (2.5,4.5) to (1.25,2);
\draw[-] (2.5,1.5) to (1.25,4);
\draw[-] (2.5,2.5) to (1.25,4);
\draw[-] (2.5,4.5) to (1.25,4);
\draw[-] (3.75,1.5) to (1.25,1);
\draw[-] (3.75,2.5) to (1.25,1);
\draw[-] (3.75,4.5) to (1.25,1);
\draw[-] (3.75,1.5) to (1.25,2);
\draw[-] (3.75,2.5) to (1.25,2);
\draw[-] (3.75,4.5) to (1.25,2);
\draw[-] (3.75,1.5) to (1.25,4);
\draw[-] (3.75,2.5) to (1.25,4);
\draw[-] (3.75,4.5) to (1.25,4);
\draw[-] (5.0,1.5) to (3.75,1);
\draw[-] (5.0,2.5) to (3.75,1);
\draw[-] (5.0,4.5) to (3.75,1);
\draw[-] (5.0,1.5) to (3.75,2);
\draw[-] (5.0,2.5) to (3.75,2);
\draw[-] (5.0,4.5) to (3.75,2);
\draw[-] (5.0,1.5) to (3.75,4);
\draw[-] (5.0,2.5) to (3.75,4);
\draw[-] (5.0,4.5) to (3.75,4);
\draw[-] (1.25,0.5) to (1.25,1);
\draw[-] (1.25,0.5) to [bend right=30] (1.25,2);
\draw[-] (1.25,0.5) to [bend right=30] (1.25,4);
\draw[-] (0.75,1) to (1.25,1);
\draw[-] (0.75,2) to (1.25,2);
\draw[-] (0.75,4) to (1.25,4);
\draw[-] (1.25,1.5) to (1.25,1);
\draw[-] (1.25,2.5) to (1.25,2);
\draw[-] (1.25,4.5) to (1.25,4);
\draw[-] (2.5,0.5) to (2.5,1);
\draw[-] (2.5,0.5) to [bend right=30] (2.5,2);
\draw[-] (2.5,0.5) to [bend right=30] (2.5,4);
\draw[-] (2,1) to (2.5,1);
\draw[-] (2,2) to (2.5,2);
\draw[-] (2,4) to (2.5,4);
\draw[-] (2.5,1.5) to (2.5,1);
\draw[-] (2.5,2.5) to (2.5,2);
\draw[-] (2.5,4.5) to (2.5,4);
\draw[-] (2.5,0.5) to (2.5,1);
\draw[-] (3.75,0.5) to (3.75,1);
\draw[-] (3.75,0.5) to [bend right=30] (3.75,2);
\draw[-] (3.75,0.5) to [bend right=30] (3.75,4);
\draw[-] (3.25,1) to (3.75,1);
\draw[-] (3.25,2) to (3.75,2);
\draw[-] (3.25,4) to (3.75,4);
\draw[-] (3.75,1.5) to (3.75,1);
\draw[-] (3.75,2.5) to (3.75,2);
\draw[-] (3.75,4.5) to (3.75,4);
\draw[-] (3.75,0.5) to (3.75,1);
\draw[-] (5.0,0.5) to (5.0,1);
\draw[-] (5.0,0.5) to [bend right=30] (5.0,2);
\draw[-] (5.0,0.5) to [bend right=30] (5.0,4);
\draw[-] (4.5,1) to (5.0,1);
\draw[-] (4.5,2) to (5.0,2);
\draw[-] (4.5,4) to (5.0,4);
\draw[-] (5.0,1.5) to (5.0,1);
\draw[-] (5.0,2.5) to (5.0,2);
\draw[-] (5.0,4.5) to (5.0,4);

\node[textnode] (l0) at (0,-0.125) {\footnotesize{\textit{turn}}};
\node[nodeknown] (p0) at (0,0.5) {};
\node[font=\tiny] (p0label) at (0,0.5) {$\lambda_{0}$};
\node[nodeunknown] (c01) at (-0.5,1) {};
\node[font=\tiny] (c01label) at (-0.5,1) {$\phi_{0_{1}}$};
\node[nodeunknown] (c02) at (-0.5,2) {};
\node[font=\tiny] (c02label) at (-0.5,2) {$\phi_{0_{2}}$};
\node[nodeunknown] (c0n) at (-0.5,4) {};
\node[font=\tiny] (c0nlabel) at (-0.5,4) {$\phi_{0_{n}}$};
\node[nodeknown] (g01) at (0,1.5) {};
\node[font=\tiny] (g01label) at (0,1.5) {$\gamma_{0_{1}}$};
\node[nodeknown] (g02) at (0,2.5) {};
\node[font=\tiny] (g02label) at (0,2.5) {$\gamma_{0_{2}}$};
\node[] (g0dots) at (0,3.375) {$\vdots$};
\node[nodeknown] (g0n) at (0,4.5) {};
\node[font=\tiny] (g0nlabel) at (0,4.5) {$\gamma_{0_{n}}$};
\node[factor] (f01) at (0,1) {};
\node[factor] (f02) at (0,2) {};
\node[factor] (f0n) at (0,4) {};
\node[nodeknown] (p1) at (1.25,0.5) {};
\node[font=\tiny] (p1label) at (1.25,0.5) {$\lambda_{1}$};
\node[nodeunknown] (c11) at (0.75,1) {};
\node[font=\tiny] (c11label) at (0.75,1) {$\phi_{1_{1}}$};
\node[nodeunknown] (c12) at (0.75,2) {};
\node[font=\tiny] (c12label) at (0.75,2) {$\phi_{1_{2}}$};
\node[nodeunknown] (c1n) at (0.75,4) {};
\node[font=\tiny] (c1nlabel) at (0.75,4) {$\phi_{1_{n}}$};
\node[nodeknown] (g11) at (1.25,1.5) {};
\node[font=\tiny] (g11label) at (1.25,1.5) {$\gamma_{1_{1}}$};
\node[nodeknown] (g12) at (1.25,2.5) {};
\node[font=\tiny] (g12label) at (1.25,2.5) {$\gamma_{1_{2}}$};
\node[] (g1dots) at (1.25,3.375) {$\vdots$};
\node[nodeknown] (g1n) at (1.25,4.5) {};
\node[font=\tiny] (g1nlabel) at (1.25,4.5) {$\gamma_{1_{n}}$};
\node[factor] (f11) at (1.25,1) {};
\node[factor] (f12) at (1.25,2) {};
\node[factor] (f1n) at (1.25,4) {};
\node[textnode] (l2) at (2.5,-0.125) {\footnotesize{\textit{the handle}}};
\node[nodeknown] (p2) at (2.5,0.5) {};
\node[font=\tiny] (p2label) at (2.5,0.5) {$\lambda_{2}$};
\node[nodeunknown] (c21) at (2,1) {};
\node[font=\tiny] (c21label) at (2,1) {$\phi_{2_{1}}$};
\node[nodeunknown] (c22) at (2,2) {};
\node[font=\tiny] (c22label) at (2,2) {$\phi_{2_{2}}$};
\node[nodeunknown] (c2n) at (2,4) {};
\node[font=\tiny] (c2nlabel) at (2,4) {$\phi_{2_{n}}$};
\node[nodeknown] (g21) at (2.5,1.5) {};
\node[font=\tiny] (g21label) at (2.5,1.5) {$\gamma_{2_{1}}$};
\node[nodeknown] (g22) at (2.5,2.5) {};
\node[font=\tiny] (g22label) at (2.5,2.5) {$\gamma_{2_{2}}$};
\node[] (g2dots) at (2.5,3.375) {$\vdots$};
\node[nodeknown] (g2n) at (2.5,4.5) {};
\node[font=\tiny] (g2nlabel) at (2.5,4.5) {$\gamma_{2_{n}}$};
\node[factor] (f21) at (2.5,1) {};
\node[factor] (f22) at (2.5,2) {};
\node[factor] (f2n) at (2.5,4) {};
\node[textnode] (l3) at (3.75,-0.125) {\footnotesize{\textit{of}}};
\node[nodeknown] (p3) at (3.75,0.5) {};
\node[font=\tiny] (p3label) at (3.75,0.5) {$\lambda_{3}$};
\node[nodeunknown] (c31) at (3.25,1) {};
\node[font=\tiny] (c31label) at (3.25,1) {$\phi_{3_{1}}$};
\node[nodeunknown] (c32) at (3.25,2) {};
\node[font=\tiny] (c32label) at (3.25,2) {$\phi_{3_{2}}$};
\node[nodeunknown] (c3n) at (3.25,4) {};
\node[font=\tiny] (c3nlabel) at (3.25,4) {$\phi_{3_{n}}$};
\node[nodeknown] (g31) at (3.75,1.5) {};
\node[font=\tiny] (g31label) at (3.75,1.5) {$\gamma_{3_{1}}$};
\node[nodeknown] (g32) at (3.75,2.5) {};
\node[font=\tiny] (g32label) at (3.75,2.5) {$\gamma_{3_{2}}$};
\node[] (g3dots) at (3.75,3.375) {$\vdots$};
\node[nodeknown] (g3n) at (3.75,4.5) {};
\node[font=\tiny] (g3nlabel) at (3.75,4.5) {$\gamma_{3_{n}}$};
\node[factor] (f31) at (3.75,1) {};
\node[factor] (f32) at (3.75,2) {};
\node[factor] (f3n) at (3.75,4) {};
\node[textnode] (l4) at (5.0,-0.125) {\footnotesize{\textit{the door}}};
\node[nodeknown] (p4) at (5.0,0.5) {};
\node[font=\tiny] (p4label) at (5.0,0.5) {$\lambda_{4}$};
\node[nodeunknown] (c41) at (4.5,1) {};
\node[font=\tiny] (c41label) at (4.5,1) {$\phi_{4_{1}}$};
\node[nodeunknown] (c42) at (4.5,2) {};
\node[font=\tiny] (c42label) at (4.5,2) {$\phi_{4_{2}}$};
\node[nodeunknown] (c4n) at (4.5,4) {};
\node[font=\tiny] (c4nlabel) at (4.5,4) {$\phi_{4_{n}}$};
\node[nodeknown] (g41) at (5.0,1.5) {};
\node[font=\tiny] (g41label) at (5.0,1.5) {$\gamma_{4_{1}}$};
\node[nodeknown] (g42) at (5.0,2.5) {};
\node[font=\tiny] (g42label) at (5.0,2.5) {$\gamma_{4_{2}}$};
\node[] (g4dots) at (5.0,3.375) {$\vdots$};
\node[nodeknown] (g4n) at (5.0,4.5) {};
\node[font=\tiny] (g4nlabel) at (5.0,4.5) {$\gamma_{4_{n}}$};
\node[factor] (f41) at (5.0,1) {};
\node[factor] (f42) at (5.0,2) {};
\node[factor] (f4n) at (5.0,4) {};
%
% PARSE TREE
%
\node[textnode] (pt1) at (0,-0.75) {\footnotesize{VB}};
\node[textnode] (pt2) at (2.25,-0.75) {\footnotesize{DT}};
\node[textnode] (pt3) at (2.75,-0.75) {\footnotesize{NN}};
\node[textnode] (pt4) at (3.75,-0.75) {\footnotesize{IN}};
\node[textnode] (pt5) at (4.75,-0.75) {\footnotesize{DT}};
\node[textnode] (pt6) at (5.25,-0.75) {\footnotesize{NN}};
\node[textnode] (pt7) at (5.0,-1.5) {\footnotesize{NP $\left(\lambda_{4}\right)$}};
\node[textnode] (pt8) at (4.375,-2.25) {\footnotesize{PP $\left(\lambda_{3}\right)$}};
\node[textnode] (pt9) at (2.5,-1.5) {\footnotesize{NP $\left(\lambda_{2}\right)$}};
\node[textnode] (pt10) at (3.4375,-3) {\footnotesize{NP $\left(\lambda_{1}\right)$}};
\node[textnode] (pt11) at (1.71875,-3.75) {\footnotesize{VP $\left(\lambda_{0}\right)$}};
\draw[] (pt2) to (2.25,-1.125) to (2.5,-1.125) to (pt9);
\draw[] (pt3) to (2.75,-1.125) to (2.5,-1.125) to (pt9);
\draw[] (pt5) to (4.75,-1.125) to (5.0,-1.125) to (pt7);
\draw[] (pt6) to (5.25,-1.125) to (5.0,-1.125) to (pt7);
\draw[] (pt4) to (3.75,-1.875) to (4.375,-1.875) to (pt8);
\draw[] (pt7) to (5.0,-1.875) to (4.375,-1.875) to (pt8);
\draw[] (pt9) to (2.5,-2.625) to (3.4375,-2.625) to (pt10);
\draw[] (pt8) to (4.375,-2.625) to (3.4375,-2.625) to (pt10);
\draw[] (pt1) to (0,-3.375) to (1.71875,-3.375) to (pt11);
\draw[] (pt10) to (3.4375,-3.375) to (1.71875,-3.375) to (pt11);
\end{tikzpicture}
	\caption{The DCG and parse tree for the expression ``turn the handle of the door''. The conditional dependence of correspondence variables $\phi_{i,j}$ on the environment model $\Upsilon$ is implicit in this visual representation of the DCG.}  
	\label{fig:turn-the-handle-of-the-door}
\end{figure}

Consider the expression ``turn the handle of the door'' that is illustrated in Figure \ref{fig:turn-the-handle-of-the-door}.  
In this example, detector inference may individually express classifiers for ``handle'' and ``door'' objects but not represent the relationship between those two objects, which may be useful for behavior planning and execution. For example, understanding the state of the door (locked, openable) and how manipulation of the door handle affects these states may rely on the represented relationship inferred by adaptive perception. 
To prevent the perception pipeline from classifying objects that might be associated with a semantic class of ``handle'' (which may include drawers, tools, cabinets, carts, etc.) to one that specializes for the door handle, the detector inference pipeline must consider a more sophisticated symbolic representation that is able to capture this relationship.  

These two examples motivate the modification of the symbolic representation of the detector class to enable adaptive perception to capture this relationship by adding the following symbol set to the space of detector symbols $\Gamma^P$ where detector types $\gamma_{t_{i}}$ and subtypes $\gamma_{st_{i}}$ are pulled from the space of semantic labels $\mathcal{S}$:
\begin{equation}
  \Gamma^{H} = \{ \gamma_{t_{i}}, \gamma_{st_{i}} | t_{i}, st_{i} \in \mathcal{S} \} 
	\label{eqn:detector_symbols}
\end{equation}

Revisiting the example in Figure \ref{fig:turn-the-handle-of-the-door}, the environment model needed for symbol grounding, behavior planning, and behavior execution will need to model both the door and door handle to localize and manipulate the handle of the door. 
While it would be possible for the behavior inference module to locate the door's handle from all handles detected by the robot, the time spent classifying and segmenting those extraneous objects could be used to facilitate other parts of a robot's intelligence architecture or increase the rate of world model updates by the perception module.  

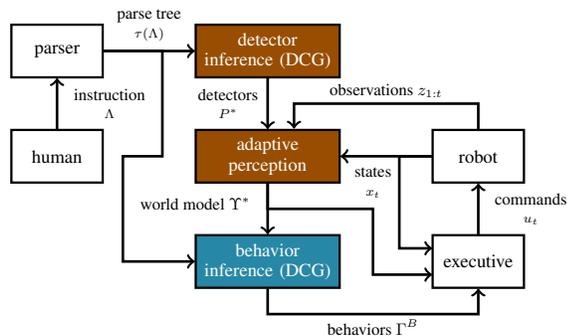
\begin{figure}[ht]
	\centering
\begin{tikzpicture}[
		ap/.style={rectangle, draw=black, text=white, fill=black!40!orange, line width=0.03cm, minimum size=10mm, text width=2.5cm, align=center},
		annotation/.style={rectangle, draw=black, text=white, fill=black!40!green, line width=0.03cm, minimum size=10mm, text width=2.5cm, align=center},
		behaviour/.style={rectangle, draw=black, text=white, fill=black!40!cyan, line width=0.03cm, minimum size=10mm, text width=2.5cm, align=center},
		robot/.style={rectangle, draw=black, fill=white!60, line width=0.03cm, minimum size=10mm, text width=1.5cm},
		% shorten >=1pt,node distance=5cm,on grid,auto
		on grid, auto,
		scale=0.7, every node/.style={transform shape},
		align=center
		]
		\node[robot]    (human) at (0,0)                            {human};
		\node[robot]    (parser)  at (0,2)                          {parser};
		\node[ap]       (nlu_pi) at (4,2)          {detector inference (DCG)};
		\node[ap]           (ap) at (4,0) {adaptive perception};
		\node[behaviour]    (nlu_bi) at (4,-2)             {behavior inference (DCG)};
		\node[robot]        (exec) at (8,-2) {executive};
		\node[robot]        (robot) at (8,0) {robot};
		\node[font=\scriptsize] (languagelabel) at (1,1) {\footnotesize instruction \\ $\Lambda$};
		\node[font=\scriptsize] (detectorslabel) at (3.25,1) {\footnotesize detectors \\ $P^*$};
		\node[font=\scriptsize] (worldlabel) at (2.625,-0.9) {\footnotesize world model $\Upsilon^*$};
		\node[font=\scriptsize] (parsetreelabel) at (1.75,2.5) {\footnotesize parse tree\\ $\tau(\Lambda)$};
		\node[font=\scriptsize] (observationslabel) at (6.25,1.25) {\footnotesize observations $z_{1:t}$};
		\node[font=\scriptsize] (stateslabel) at (6,-0.5) {\footnotesize states  \\ $x_t$};
		\node[font=\scriptsize] (behaviorslabel) at (6,-3.25) {\footnotesize behaviors $\Gamma^{B}$};
		\node[font=\scriptsize] (commandslabel) at (9,-1) {\footnotesize commands \\ $u_t$};
		\draw [->,line width=1pt] (human.north) -- (parser.south);
		\draw [->,line width=1pt] (parser.east) -- (nlu_pi.west);
		\draw [->,line width=1pt] (nlu_pi.south) -- (ap.north);
		\draw [->,line width=1pt] (ap.south) -- (nlu_bi.north);
		\draw [->,line width=1pt] (nlu_bi.south) -- (4,-3) -- (8,-3) -- (exec.south);
		\draw [->,line width=1pt] (robot.west) -- (ap.east);
		\draw [->,line width=1pt] (exec.north) -- (robot.south);
		\draw [->,line width=1pt] (ap.south) -- (4,-1) -- (6,-1) -- (6,-2.25) -- (7.125,-2.25);
		\draw [->,line width=1pt] (robot.west) -- (6.5,0) -- (6.5,-1.75) -- (7.125,-1.75);
		\draw [->,line width=1pt] (robot.north) -- (8,1) -- (4.5,1) -- (4.5,0.525);
		\draw [->,line width=1pt] (parser.east) -- (2,2) -- (2,0) -- (1.25,0) -- (1.25,-2) -- (nlu_bi.west);
		\end{tikzpicture}
	\caption{The system architecture used to perform the mobile manipulation experiments.}
	\label{fig:system-architecture}
\end{figure}

Figure \ref{fig:system-architecture} provides an overview of the system architecture. Tasks begin when an instruction is provided to the robot. The instruction is parsed and passed to the DCG-based detector inference node to infer the minimal set of classifiers $P^{*}$ needed to understand, plan, and execute the behavior. The adaptive perception model uses these classifiers to construct minimal world models $\Upsilon^{*}$ for DCG-based behavior inference and the executive module to plan and execute these behaviors.  
	
\begin{figure*}
	\centering
	\begin{tikzpicture}[
		function/.style={rectangle, draw=black, fill=white!60, line width=0.03cm, minimum size=10mm},
		decision/.style={draw=black, diamond, aspect=2, fill=white!60, line width=0.03cm, minimum size=6mm},
		enter/.style={circle,draw=black, aspect=2, fill=white!60, line width=0.03cm, minimum size=6mm},
		on grid, auto,
		scale=0.7, every node/.style={transform shape},
		align=center
		]
	
		\node[enter, text width=3cm]			(rx)						{RECEIVE\\ BEHAVIOR(A,B)};	
		\node[decision]		(isNav) [right=1.5in of rx]				{is nav};
		\node[decision]		(isOpen) [below=1in of isNav]			{is open};
		\node[function]		(planA) [right=1.3in of isOpen]				{NAVIGATE(A)};
		\node[function]		(planA_top) [right=1.3in of isNav]				{NAVIGATE(A)};
		\node[decision]		(detect)		[right=1.3in of planA]				{DETECT(B)};
		\node[function]		(localize)	[right=1.5in of detect]				{LOCALIZE(B)};
		\node[function]		(turn)			[right=1in of localize]				{TURN(B)};
		\node[function]		(push)			[right=0.85in of turn]					{PUSH(A)};
		
		\node[enter]			(failure)		[below=1in of detect]					{FAILURE};
		\node[enter]			(failure2)		[below=1in of isOpen]					{FAILURE};
	
		\node[enter]			(completeA)	[right=1.2in of planA_top]			{COMPLETE};
		\node[enter]			(completeB)	[right=1in of push]					{COMPLETE};
	
		\draw [->, line width=1pt] (rx.east) -- (isNav.west);
	
		\draw [->, line width=1pt] (isNav.south)  -- node [left] {\footnotesize no} (isOpen.north);
		\draw [->, line width=1pt] (isNav.east)  -- node [below] {\footnotesize yes} (planA_top.west);
		\draw [->, line width=1pt] (isOpen.east)  -- node [below] {\footnotesize yes} (planA.west);
	
		\draw [->, line width=1pt]	(planA.east) -- (detect.west);
		\draw [->, line width=1pt]  (detect.east) -- node [above] {\footnotesize yes} (localize.west);
		\draw [->, line width=1pt]	(localize.east) -- (turn.west);
		\draw [->, line width=1pt]  (turn.east) -- (push.west);
		\draw [->, line width=1pt]  (detect.south) -- node [left] {\footnotesize no} (failure.north);
		
		\draw [->, line width=1pt]  (planA_top.east) -- (completeA.west);
		\draw [->, line width=1pt]  (push.east) -- (completeB.west);
	
		\draw [->, line width=1pt]  (isOpen.south) -- node [left] {\footnotesize no} (failure2.north);
		\end{tikzpicture}
	\caption{The flow chart describing the operation of the executive node in Figure \ref{fig:system-architecture}. The functions are defined as follows. NAVIGATE(A) moves the robot's base to a region with fixed displacement from A. DETECT(B) checks the current world model for the presence of an object with type B. LOCALIZE(B) moves the manipulator and uses force feedback to localize object B. TURN(B) moves the manipulator to rotate object B until a torque threshold is exceeded. Finally, PUSH(A) uses the manipulator to displace object A a fixed distance. With the ``drive to the door'' instruction A would be the door, and B would have no value. For ``open the door'', A would again be the door and B would represent the door handle.}
	\label{fig:state-machines}
	\end{figure*}
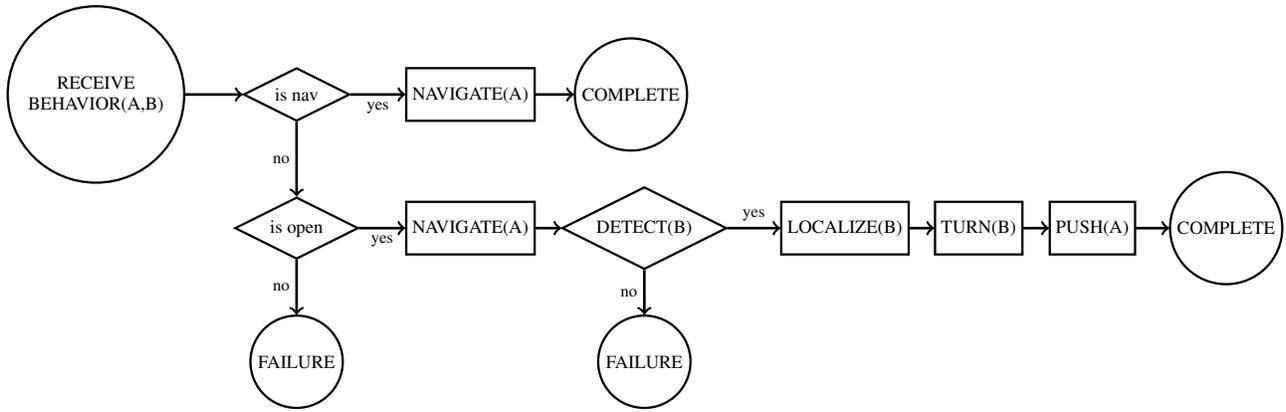

The parsed instruction and inferred environment representation are passed to the DCG-based natural language understanding node used for behavior inference as previously described in \cite{patki2019a}. The executive node shown in Figure \ref{fig:system-architecture} acts as a state machine to govern the robot's behaviors during the task. After receiving behavior symbols from behavior inference, it sends requests to the arm and base planner based on the world that is being continuously updated by adaptive perception. The executive node follows the flow chart shown in Figure \ref{fig:state-machines}.

\section{Experimental Setup}

\begin{figure}[ht]
  \centering
  \includegraphics[width=0.9\columnwidth]{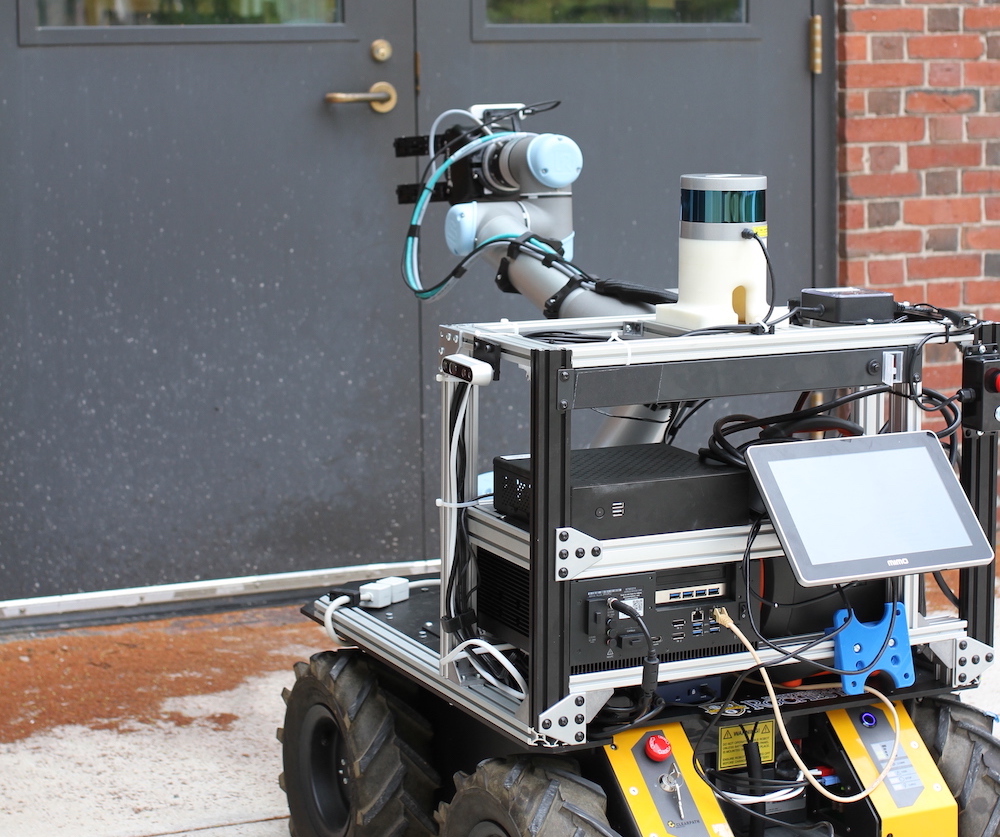} 
  \caption{The mobile manipulation platform and sensor stack used for experimentation.}
  \label{fig:experimental_platform}
\end{figure}

Experiments are performed on the platform shown in Figure \ref{fig:experimental_platform}: a Clearpath Robotics A200 Husky with a mounted Universal Robotics UR5 arm and a Robotiq 3-finger Adaptive Robot Gripper. A Robotiq FT300 force torque sensor is placed between the arm and gripper. We use a Velodyne VLP-16 lidar with Cartographer \cite{googleCartographer} to perform planar Simultaneous Localization and Mapping (SLAM). An Intel RealSense D435 depth camera is mounted on the end effector and is used for object detection. Detectors in the perception pipeline are trained YOLOv3 networks \cite{yolov3} to identify and generate bounding boxes around objects in the RGB images captured from the RealSense. These bounding boxes are used to segment the depth data provided by the RealSense to generate 3D bounding boxes. These objects are added with their bounding boxes to a global world model using the odometry output by SLAM. 

The arm planner uses trac\_ik \cite{Beeson-humanoids-15} to perform inverse kinematics to find joint angles that satisfy pose constraints of the manipulator. A trajectory is then generated to linearly transition each joint between its current pose and the inverse kinematics solution. A simple pure-pursuit style controller \cite{Coulter-1992-13338} is used for the NAVIGATE function in Figure \ref{fig:state-machines} to drive the robot between poses. Inspired by \cite{nextBest} we explore deliberative interactive estimation to turn the door handle. The LOCALIZE and TURN functions in Figure \ref{fig:state-machines} use the force torque sensor to assist in localizing the door handle and to turn it to a sufficient degree. LOCALIZE first moves the arm to contact with the door above the visually localized door handle. The arm then moves down until contact is detected with the door handle. TURN then continues this downward motion until a torque limit is exceeded indicating the door handle has reached the end of its range of motion. The door is then pushed open.

The behavior inference corpus was trained with 115 annotated examples. Each example contains an annotated parse tree of the natural language instruction, a world model comprised of objects, and the correct grounding symbols for the world parse-tree pair. A small focused corpus of 7 annotated examples were used to train the Adaptive Perception NLU in Figure \ref{fig:system-architecture}. These annotated examples take the same form as those for behavior inference but do not contain a world model, as the perception-symbol grounding is world model agnostic. The generalizability and ability of DCG to ground natural language commands from more diverse corpora has been demonstrated in previous work \cite{paul16a,paul2018efficient}. These examples were engineered to explore the examples studied in this paper using grammatical patterns of verbs, prepositions, and noun phrases observed in the natural language corpora described in \cite{paul2018efficient}. A lexicon of the adaptive perception NLU corpus is provided below.

\begin{flushleft}
	\noindent
	\begin{flalign*}
	& VB \rightarrow \{drive|open|turn|look\} \\
	& DT \rightarrow \{the\} \\
	& NN \rightarrow \{door|handle|drawer|box|top\} \\
	& IN \rightarrow \{through|of\} \\
\end{flalign*}
\end{flushleft}
	
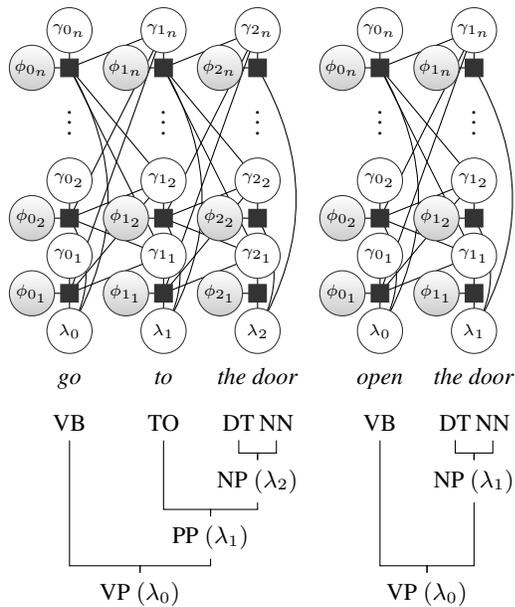
\begin{figure}[htb]
\centering
\begin{tikzpicture}[textnode/.style={anchor=mid,font=\tiny},nodeknown/.style={circle,draw=black!80,fill=white,minimum size=6mm,font=\tiny},nodeunknown/.style={circle,draw=black!80,fill=black!10,minimum size=6mm,font=\tiny,top color=white,bottom color=black!20},factor/.style={rectangle,draw=black!80,fill=black!80,minimum size=2mm,font=\tiny,text=white}]
\draw[-] (0,0.5) to (0,1);
\draw[-] (0,0.5) to [bend right=30] (0,2);
\draw[-] (0,0.5) to [bend right=30] (0,4);
\draw[-] (-0.5,1) to (0,1);
\draw[-] (-0.5,2) to (0,2);
\draw[-] (-0.5,4) to (0,4);
\draw[-] (0,1.5) to (0,1);
\draw[-] (0,2.5) to (0,2);
\draw[-] (0,4.5) to (0,4);
\draw[-] (1.25,1.5) to (0,1);
\draw[-] (1.25,2.5) to (0,1);
\draw[-] (1.25,4.5) to (0,1);
\draw[-] (1.25,1.5) to (0,2);
\draw[-] (1.25,2.5) to (0,2);
\draw[-] (1.25,4.5) to (0,2);
\draw[-] (1.25,1.5) to (0,4);
\draw[-] (1.25,2.5) to (0,4);
\draw[-] (1.25,4.5) to (0,4);
\draw[-] (2.5,1.5) to (1.25,1);
\draw[-] (2.5,2.5) to (1.25,1);
\draw[-] (2.5,4.5) to (1.25,1);
\draw[-] (2.5,1.5) to (1.25,2);
\draw[-] (2.5,2.5) to (1.25,2);
\draw[-] (2.5,4.5) to (1.25,2);
\draw[-] (2.5,1.5) to (1.25,4);
\draw[-] (2.5,2.5) to (1.25,4);
\draw[-] (2.5,4.5) to (1.25,4);
\draw[-] (1.25,0.5) to (1.25,1);
\draw[-] (1.25,0.5) to [bend right=30] (1.25,2);
\draw[-] (1.25,0.5) to [bend right=30] (1.25,4);
\draw[-] (0.75,1) to (1.25,1);
\draw[-] (0.75,2) to (1.25,2);
\draw[-] (0.75,4) to (1.25,4);
\draw[-] (1.25,1.5) to (1.25,1);
\draw[-] (1.25,2.5) to (1.25,2);
\draw[-] (1.25,4.5) to (1.25,4);
\draw[-] (2.5,0.5) to (2.5,1);
\draw[-] (2.5,0.5) to [bend right=30] (2.5,2);
\draw[-] (2.5,0.5) to [bend right=30] (2.5,4);
\draw[-] (2,1) to (2.5,1);
\draw[-] (2,2) to (2.5,2);
\draw[-] (2,4) to (2.5,4);
\draw[-] (2.5,1.5) to (2.5,1);
\draw[-] (2.5,2.5) to (2.5,2);
\draw[-] (2.5,4.5) to (2.5,4);

\node[textnode] (l0) at (0,-0.125) {\footnotesize{\textit{go}}};
\node[nodeknown] (p0) at (0,0.5) {};
\node[font=\tiny] (p0label) at (0,0.5) {$\lambda_{0}$};
\node[nodeunknown] (c01) at (-0.5,1) {};
\node[font=\tiny] (c01label) at (-0.5,1) {$\phi_{0_{1}}$};
\node[nodeunknown] (c02) at (-0.5,2) {};
\node[font=\tiny] (c02label) at (-0.5,2) {$\phi_{0_{2}}$};
\node[nodeunknown] (c0n) at (-0.5,4) {};
\node[font=\tiny] (c0nlabel) at (-0.5,4) {$\phi_{0_{n}}$};
\node[nodeknown] (g01) at (0,1.5) {};
\node[font=\tiny] (g01label) at (0,1.5) {$\gamma_{0_{1}}$};
\node[nodeknown] (g02) at (0,2.5) {};
\node[font=\tiny] (g02label) at (0,2.5) {$\gamma_{0_{2}}$};
\node[] (g0dots) at (0,3.375) {$\vdots$};
\node[nodeknown] (g0n) at (0,4.5) {};
\node[font=\tiny] (g0nlabel) at (0,4.5) {$\gamma_{0_{n}}$};
\node[factor] (f01) at (0,1) {};
\node[factor] (f02) at (0,2) {};
\node[factor] (f0n) at (0,4) {};
\node[textnode] (l1) at (1.25,-0.125) {\footnotesize{\textit{to}}};
\node[nodeknown] (p1) at (1.25,0.5) {};
\node[font=\tiny] (p1label) at (1.25,0.5) {$\lambda_{1}$};
\node[nodeunknown] (c11) at (0.75,1) {};
\node[font=\tiny] (c11label) at (0.75,1) {$\phi_{1_{1}}$};
\node[nodeunknown] (c12) at (0.75,2) {};
\node[font=\tiny] (c12label) at (0.75,2) {$\phi_{1_{2}}$};
\node[nodeunknown] (c1n) at (0.75,4) {};
\node[font=\tiny] (c1nlabel) at (0.75,4) {$\phi_{1_{n}}$};
\node[nodeknown] (g11) at (1.25,1.5) {};
\node[font=\tiny] (g11label) at (1.25,1.5) {$\gamma_{1_{1}}$};
\node[nodeknown] (g12) at (1.25,2.5) {};
\node[font=\tiny] (g12label) at (1.25,2.5) {$\gamma_{1_{2}}$};
\node[] (g1dots) at (1.25,3.375) {$\vdots$};
\node[nodeknown] (g1n) at (1.25,4.5) {};
\node[font=\tiny] (g1nlabel) at (1.25,4.5) {$\gamma_{1_{n}}$};
\node[factor] (f11) at (1.25,1) {};
\node[factor] (f12) at (1.25,2) {};
\node[factor] (f1n) at (1.25,4) {};
\node[textnode] (l2) at (2.5,-0.125) {\footnotesize{\textit{the door}}};
\node[nodeknown] (p2) at (2.5,0.5) {};
\node[font=\tiny] (p2label) at (2.5,0.5) {$\lambda_{2}$};
\node[nodeunknown] (c21) at (2,1) {};
\node[font=\tiny] (c21label) at (2,1) {$\phi_{2_{1}}$};
\node[nodeunknown] (c22) at (2,2) {};
\node[font=\tiny] (c22label) at (2,2) {$\phi_{2_{2}}$};
\node[nodeunknown] (c2n) at (2,4) {};
\node[font=\tiny] (c2nlabel) at (2,4) {$\phi_{2_{n}}$};
\node[nodeknown] (g21) at (2.5,1.5) {};
\node[font=\tiny] (g21label) at (2.5,1.5) {$\gamma_{2_{1}}$};
\node[nodeknown] (g22) at (2.5,2.5) {};
\node[font=\tiny] (g22label) at (2.5,2.5) {$\gamma_{2_{2}}$};
\node[] (g2dots) at (2.5,3.375) {$\vdots$};
\node[nodeknown] (g2n) at (2.5,4.5) {};
\node[font=\tiny] (g2nlabel) at (2.5,4.5) {$\gamma_{2_{n}}$};
\node[factor] (f21) at (2.5,1) {};
\node[factor] (f22) at (2.5,2) {};
\node[factor] (f2n) at (2.5,4) {};
%
% PARSE TREE
%
\node[textnode] (pt1) at (0,-0.75) {\footnotesize{VB}};
\node[textnode] (pt2) at (1.25,-0.75) {\footnotesize{TO}};
\node[textnode] (pt3) at (2.25,-0.75) {\footnotesize{DT}};
\node[textnode] (pt4) at (2.75,-0.75) {\footnotesize{NN}};
\node[textnode] (pt5) at (2.5,-1.5) {\footnotesize{NP $\left(\lambda_{2}\right)$}};
\node[textnode] (pt6) at (1.875,-2.25) {\footnotesize{PP $\left(\lambda_{1}\right)$}};
\node[textnode] (pt7) at (0.9375,-3) {\footnotesize{VP $\left(\lambda_{0}\right)$}};
\draw[] (pt3) to (2.25,-1.125) to (2.5,-1.125) to (pt5);
\draw[] (pt4) to (2.75,-1.125) to (2.5,-1.125) to (pt5);
\draw[] (pt2) to (1.25,-1.875) to (1.875,-1.875) to (pt6);
\draw[] (pt5) to (2.5,-1.875) to (1.875,-1.875) to (pt6);
\draw[] (pt1) to (0,-2.625) to (0.9375,-2.625) to (pt7);
\draw[] (pt6) to (1.875,-2.625) to (0.9375,-2.625) to (pt7);
\end{tikzpicture}
\begin{tikzpicture}[textnode/.style={anchor=mid,font=\tiny},nodeknown/.style={circle,draw=black!80,fill=white,minimum size=6mm,font=\tiny},nodeunknown/.style={circle,draw=black!80,fill=black!10,minimum size=6mm,font=\tiny,top color=white,bottom color=black!20},factor/.style={rectangle,draw=black!80,fill=black!80,minimum size=2mm,font=\tiny,text=white}]
\draw[-] (0,0.5) to (0,1);
\draw[-] (0,0.5) to [bend right=30] (0,2);
\draw[-] (0,0.5) to [bend right=30] (0,4);
\draw[-] (-0.5,1) to (0,1);
\draw[-] (-0.5,2) to (0,2);
\draw[-] (-0.5,4) to (0,4);
\draw[-] (0,1.5) to (0,1);
\draw[-] (0,2.5) to (0,2);
\draw[-] (0,4.5) to (0,4);
\draw[-] (1.25,1.5) to (0,1);
\draw[-] (1.25,2.5) to (0,1);
\draw[-] (1.25,4.5) to (0,1);
\draw[-] (1.25,1.5) to (0,2);
\draw[-] (1.25,2.5) to (0,2);
\draw[-] (1.25,4.5) to (0,2);
\draw[-] (1.25,1.5) to (0,4);
\draw[-] (1.25,2.5) to (0,4);
\draw[-] (1.25,4.5) to (0,4);
\draw[-] (1.25,0.5) to (1.25,1);
\draw[-] (1.25,0.5) to [bend right=30] (1.25,2);
\draw[-] (1.25,0.5) to [bend right=30] (1.25,4);
\draw[-] (0.75,1) to (1.25,1);
\draw[-] (0.75,2) to (1.25,2);
\draw[-] (0.75,4) to (1.25,4);
\draw[-] (1.25,1.5) to (1.25,1);
\draw[-] (1.25,2.5) to (1.25,2);
\draw[-] (1.25,4.5) to (1.25,4);
\node[textnode] (l0) at (0,-0.125) {\footnotesize{\textit{open}}};
\node[nodeknown] (p0) at (0,0.5) {};
\node[font=\tiny] (p0label) at (0,0.5) {$\lambda_{0}$};
\node[nodeunknown] (c01) at (-0.5,1) {};
\node[font=\tiny] (c01label) at (-0.5,1) {$\phi_{0_{1}}$};
\node[nodeunknown] (c02) at (-0.5,2) {};
\node[font=\tiny] (c02label) at (-0.5,2) {$\phi_{0_{2}}$};
\node[nodeunknown] (c0n) at (-0.5,4) {};
\node[font=\tiny] (c0nlabel) at (-0.5,4) {$\phi_{0_{n}}$};
\node[nodeknown] (g01) at (0,1.5) {};
\node[font=\tiny] (g01label) at (0,1.5) {$\gamma_{0_{1}}$};
\node[nodeknown] (g02) at (0,2.5) {};
\node[font=\tiny] (g02label) at (0,2.5) {$\gamma_{0_{2}}$};
\node[] (g0dots) at (0,3.375) {$\vdots$};
\node[nodeknown] (g0n) at (0,4.5) {};
\node[font=\tiny] (g0nlabel) at (0,4.5) {$\gamma_{0_{n}}$};
\node[factor] (f01) at (0,1) {};
\node[factor] (f02) at (0,2) {};
\node[factor] (f0n) at (0,4) {};
\node[textnode] (l1) at (1.25,-0.125) {\footnotesize{\textit{the door}}};
\node[nodeknown] (p1) at (1.25,0.5) {};
\node[font=\tiny] (p1label) at (1.25,0.5) {$\lambda_{1}$};
\node[nodeunknown] (c11) at (0.75,1) {};
\node[font=\tiny] (c11label) at (0.75,1) {$\phi_{1_{1}}$};
\node[nodeunknown] (c12) at (0.75,2) {};
\node[font=\tiny] (c12label) at (0.75,2) {$\phi_{1_{2}}$};
\node[nodeunknown] (c1n) at (0.75,4) {};
\node[font=\tiny] (c1nlabel) at (0.75,4) {$\phi_{1_{n}}$};
\node[nodeknown] (g11) at (1.25,1.5) {};
\node[font=\tiny] (g11label) at (1.25,1.5) {$\gamma_{1_{1}}$};
\node[nodeknown] (g12) at (1.25,2.5) {};
\node[font=\tiny] (g12label) at (1.25,2.5) {$\gamma_{1_{2}}$};
\node[] (g1dots) at (1.25,3.375) {$\vdots$};
\node[nodeknown] (g1n) at (1.25,4.5) {};
\node[font=\tiny] (g1nlabel) at (1.25,4.5) {$\gamma_{1_{n}}$};
\node[factor] (f11) at (1.25,1) {};
\node[factor] (f12) at (1.25,2) {};
\node[factor] (f1n) at (1.25,4) {};
%
% PARSE TREE
%
\node[textnode] (pt1) at (0,-0.75) {\footnotesize{VB}};
\node[textnode] (pt2) at (1.0,-0.75) {\footnotesize{DT}};
\node[textnode] (pt3) at (1.5,-0.75) {\footnotesize{NN}};
\node[textnode] (pt4) at (1.25,-1.5) {\footnotesize{NP $\left(\lambda_{1}\right)$}};
\node[textnode] (pt5) at (0.625,-3) {\footnotesize{VP $\left(\lambda_{0}\right)$}};
\draw[] (pt2) to (1.0,-1.125) to (1.25,-1.125) to (pt4);
\draw[] (pt3) to (1.5,-1.125) to (1.25,-1.125) to (pt4);
\draw[] (pt1) to (0,-2.625) to (0.625,-2.625) to (pt5);
\draw[] (pt4) to (1.25,-2.625) to (0.625,-2.625) to (pt5);
\end{tikzpicture}
\caption{Examples of DCGs and parse trees for the instructions described in the physical experiments. The adaptive perception DCGs for ``go to the door'' expresses a symbol for a door detector while the adaptive perception DCG for ``open the door'' infers symbols for both the door and door handle detector, as well as the hierarchy between them. The behavior inference DCG for ``go to the door'' generates a ``navigate'' action to a specific door object while the behavior inference DCG for ``open the door'' expresses an ``open'' action to the same door. While the symbol for behavior inference does not need the door handle object to be detected, the state machine that implements the ``open'' behavior requires a handle to complete some state transitions.}
\label{fig:instructions}
\end{figure}

To evaluate the hierarchical perception symbols, the robot was provided with two different tasks requiring it to decompose the same object in different ways. The two instructions prompting the tasks were ``open the door'' and ``drive to the door'' as illustrated in Figure \ref{fig:instructions}. In these experiments we send parse trees of instructions to the NLU nodes. In the following section we present images of these experiments, the world models generated by the robot for each task, and a measure of perception time for each experiment. To demonstrate the cost of generating exhaustive world models we provide a baseline with a static number of detectors (5) in one trial with the task ``drive to the door''. A single trial was completed for each command to explore the functionality of grounding and execution; a through evaluation of hierarchical symbolic representations in adaptive perception is left for future work.

\section{Results}
Table \ref{table:percp-rates} demonstrates the advantage of building adaptive worlds through the order of magnitude difference between the exhaustive perception (EP) baseline and the adaptive trials. This matches the observations described in previous publications on adaptive perception \cite{patki18a,patki2019a}, which is that adaptive representations scale with an increase in the number of detectors, while exhaustive representations do not. The average perception loop period for the ``open the door'' task is seen to be roughly twice that of the ``drive to the door'' AP task. This corresponds with the two detectors inferred to be necessary for the opening task as opposed to the one detector inferred to be necessary for the drive to task. The active detectors in each task is provided as well in Table \ref{table:percp-rates}.

\begin{table}[htb]
	\centering
	\begin{tabular}{P{0.3\columnwidth}P{0.26\columnwidth}P{0.28\columnwidth}}
	\toprule
	instruction & avg. perception period (s) & active detectors \\
	\midrule
	drive to the door (EP) & 2.060 & pitcher, cracker box, door, ball, suitcase \\
	drive to the door (AP) & 0.092 & door \\   
	open the door (AP) & 0.158 & door, door handle  \\
	\bottomrule      
	\end{tabular}
	\caption{Average perception periods and active detectors during the experiments where the natural language commands ``drive to the door'' and ``open the door'' were issued.}
	\label{table:percp-rates}
\end{table}

\begin{figure*}[htb]
	\centering
	\mbox{
	\begin{subfigure}[b]{0.95\textwidth}
		\centering
		\includegraphics[width=\textwidth]{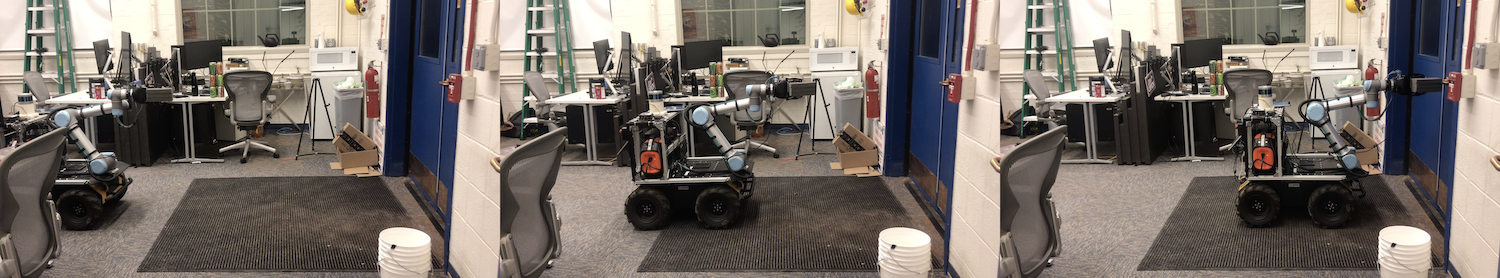}
		\caption{The Husky A200 UGV performing the task inferred from ``drive to the door''}
	\end{subfigure}
	} \\
	\mbox{
	\begin{subfigure}[b]{0.95\textwidth}
		\centering
		\includegraphics[width=\textwidth]{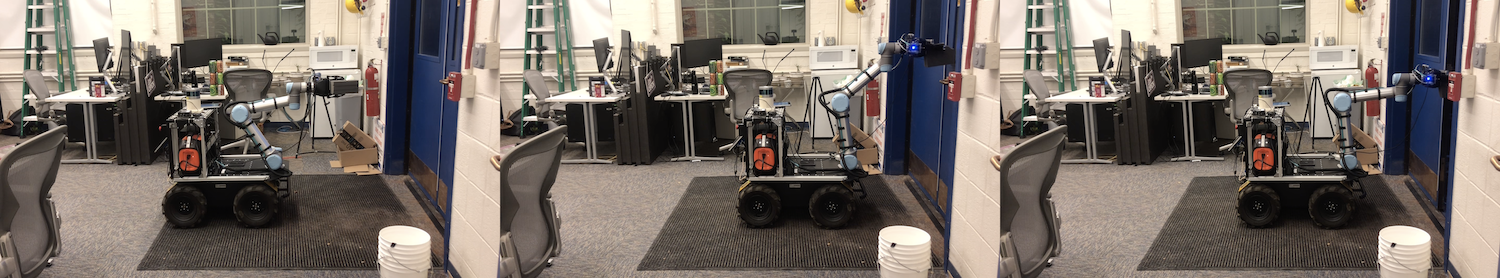}
		\caption{The Husky A200 UGV performing the task inferred from ``open the door''}
	\end{subfigure}
	}
  \caption{A sequence of images depicting the course of events for each experiment. The set of the images on the top show the robot navigating to the door after receiving the command ``drive to the door''. The bottom set of images show the robot navigating to and opening the door after receiving the command ``open the door''.}
  \label{fig:film-strip}
\end{figure*}

Figure \ref{fig:film-strip} depicts the robot's actions as it completes each commanded task. The top row shows the robot navigating from its initial position in the leftmost image to close proximity to the detected door in the rightmost image when provided the instruction ``drive to the door''. The bottom row depicts the completion of the natural language instruction ``open the door''. In the leftmost panel the robot is seen navigating to the door. The center panel depicts the LOCALIZE function in Figure \ref{fig:state-machines} where the bottom of the gripper makes contact with the door and door handle objects, and the rightmost panel depicts the robot after the completion of the PUSH function.

\begin{figure}[htb]
	\centering
	\begin{subfigure}[b]{0.9\columnwidth}
		\centering
		\includegraphics[width=\textwidth]{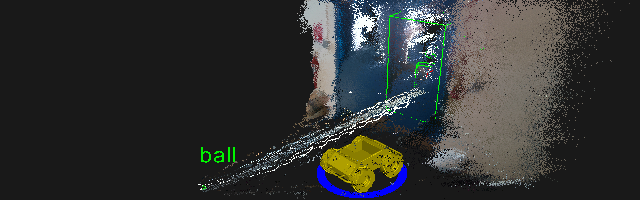}
		\caption{``drive to the door'' with the exhaustive baseline}
	\end{subfigure} \\
	\begin{subfigure}[b]{0.9\columnwidth}
		\centering
		\includegraphics[width=\textwidth]{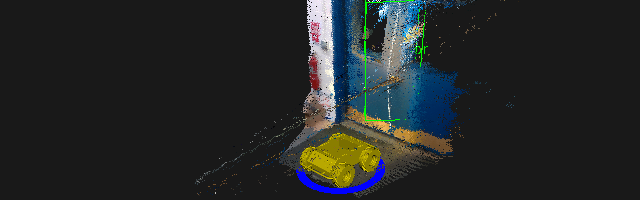}
		\caption{``drive to the door'' with adaptive perception}
	\end{subfigure} \\
	\begin{subfigure}[b]{0.9\columnwidth}
		\centering
		\includegraphics[width=\textwidth]{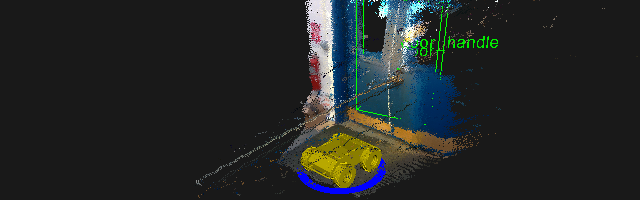}
		\caption{``open the door'' with adaptive perception}
	\end{subfigure}
  \caption{The robot's world model at the end of each experiment. The top and middle images are the result of the command ``drive to the door'' where the top image is the result of the exhaustive baseline, and the middle image is the result of adaptive perception for the same task. The bottom image is the world model generated by adaptive perception after completing the task ``open the door''. }
  \label{fig:vis-view}
\end{figure}

The world models in Figure \ref{fig:vis-view} illustrate the robot's world representation after the completion of each task under different perception configurations. In the top image, generated while completing the ``drive to the door'' instruction while running the baseline exhaustive pipeline, the presence of false detections of irrelevant objects can be seen close to the robot's initial position (ball) and below the window of the door. In the middle image, generated by adaptive perception for the same instruction, the only detected object was the door. In the bottom image, adaptive perception detects only the door and door handle needed to complete the behavior inferred for the instruction ``open the door''.

\section{Discussion}
As shown in the previous section, we have demonstrated that adaptive perception has the ability to infer hierarchical perceptual classifier symbols not explicitly expressed in the command. When tasked with ``drive to the door'', adaptive perception created a world model with just a door, whereas when tasked with ``open the door'' adaptive perception inferred the importance of decomposing the door into the door and its handle, even though the latter object was not mentioned explicitly in the utterance. It also represented the hierarchy between the object detectors for these two objects. These objects were then used to complete the task commanded as can be seen in Figure \ref{fig:film-strip}. 

The demonstration in this paper takes a first pass at simplifying world representations for language understanding and planning by considering object hierarchies; however, other factors not addressed in this work are necessary to converge on a truly minimal viable world representation. This paper does not consider inferring object fidelity from natural language, and world models are constructed of only static rigid bodies. The planning done to complete these experiments makes assumptions about the dynamics of the door that are not explicitly modeled by the robot.

The perception classifier symbolic hierarchies introduced in this paper only have one level of decomposition and do not represent hierarchies with additional layers. In future work we plan to expand the proposed hierarchical representation and incorporate the modeling of fidelity with planning requirements inferred from natural language. This demonstration would benefit from more thorough experimental evaluation on a wider range of planning and manipulation tasks in different environments. 

Another area of exploration involves exploiting information contained in the hierarchical representation of inferred detectors to build more robust and filtered perception pipelines. Consider the task ``remove the bolt from the chair's arm rest''. Our proposed perception symbols would represent the hierarchy between the arm rest and the bolt, but the bolt classifier would likely not only classify bolts in the context of arm rests. By filtering observations by proximity to parent object detections, or by only enabling detectors when parent objects are detected, world models can further approach their minimally viable forms. This filtering would also increase the robustness of a perception pipeline as false detections could be removed if they do not appear in the context of parent objects. This object filtering becomes even more important as experiments and the hierarchies encountered scale up. For a long-term navigation task where robots traverse between multiple buildings the likelihood of false detections increases with the duration and movement of the robot. Limiting the use of these detections to areas where they are consistent with the explicit hierarchical context relevant to the task at hand is expected to facilitate the process of building accurate and scalable spatial-semantic-metric representations of the environment. 
Development of behavior planners that reason over the hierarchies produced by adaptive perception would also be important to experimentally validate the utility of such representations.

\section{Conclusions}
This paper has explored a novel symbolic representation for language-guided adaptive perception where relationships between object detectors are inferred. The value of adapting perception pipelines to consider the relationship and fidelity of object models improves a robot's ability to move from using objects as mere landmarks in tasks to ones that require interaction and manipulation of object states. These examples exhibit similar improvements over exhaustive baselines that are observed in \cite{patki18a,patki2019a} and show how detector classes could be influenced by verb phrases using synthetic corpora engineered to explore language instructions for different tasks involving the same objects. We have demonstrated on a physical system adaptive perception's ability to construct a minimal viable world to complete tasks in which the hierarchies of detectors are not explicit in the instructions given. 

The ideas and observations described in this paper lead to other avenues of future work in grounded language communication. In addition to exploring richer corpora composed of more diverse language, detectors, environments, and behaviors, expanding the symbolic representation to more specifically describe the fidelity of motion models for detected objects would expand our ability to plan interactions with objects without a priori assumptions in the motion planning framework. 

\section{Acknowledgments}
This work was supported in part by the National Science Foundation under grant IIS-1637813.

\bibliographystyle{aaai}
\bibliography{fahnestock-aaai-ai-hri-2019.bib}

\end{document}